\documentclass{article}

\PassOptionsToPackage{numbers, sort&compress}{natbib}
\usepackage[preprint]{main}

\usepackage[utf8]{inputenc} 
\usepackage[T1]{fontenc}    
\usepackage{url}            
\usepackage{booktabs}       
\usepackage{amsfonts}       
\usepackage{nicefrac}       
\usepackage{microtype}      
\usepackage{enumitem}
\usepackage{wrapfig} 
\usepackage{graphicx}
\usepackage{subfigure}
\usepackage{amsmath}
\usepackage{amssymb}
\usepackage{booktabs}
\usepackage{bbm}
\usepackage[dvipsnames,svgnames,x11names,table]{xcolor} 
\usepackage{pifont}
\usepackage{bbding}

\usepackage{wrapfig}
\usepackage{multirow}
\usepackage{makecell}
\usepackage{transparent}
\usepackage{tikz}
\usepackage{mathrsfs}
\usepackage{dutchcal}
\usepackage{color}
\usepackage[table]{xcolor}

\definecolor{pretty-blue}{RGB}{0, 113, 188}
\definecolor{brown}{RGB}{201, 104, 71}

\usepackage[colorlinks=True,
            linkcolor=blue,
            anchorcolor=blue,  
            pagebackref,
            urlcolor=blue,
            citecolor=teal,
            ]{hyperref}

\title{\ \ \ \ \ \ ${\mathscr{F}\mathcal{ox}}$: Focus Anywhere for Fine-grained\\ \  \ \ \ \ \ \ \ Multi-page Document Understanding }

\author{
Chenglong Liu$^{1}$~\thanks{This work was done when the first author was interning at Megvii Technology Inc.}, Haoran Wei$^{2}$~,
Jinyue Chen$^{1}$, Lingyu Kong$^{1}$,   \bf{Zheng Ge}$^{2}$~, \\ \bf{Zining Zhu}$^{1}$, \bf{Liang Zhao}$^{2}$, \bf{Jianjian Sun}$^{2}$, \bf{Chunrui Han}$^{2}$, Xiangyu Zhang$^{2}$\\
 \ \  $^{1}$University of Chinese Academy of Sciences, $^{2}$MEGVII Technology  \\ 
{\url{https://ucaslcl.github.io/foxhome/}}
}

\begin{document}

\maketitle

\begin{tikzpicture}[remember picture,overlay,shift={(current page.north west)}]
\node[anchor=north west, xshift=3.6cm, yshift=-2.8cm]{\scalebox{1}[1]{\includegraphics[width=2.6cm]{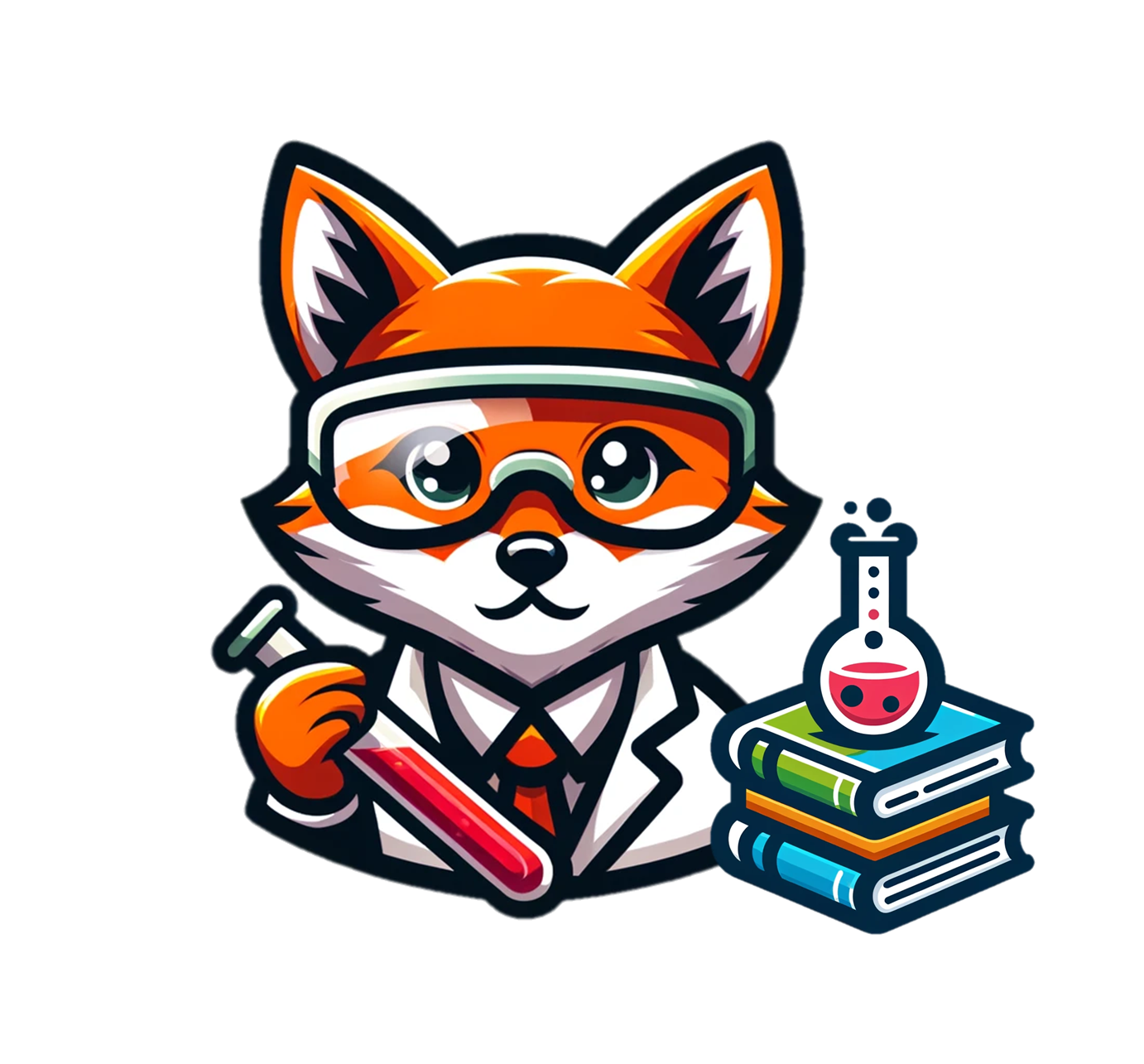}}};
\end{tikzpicture}




\begin{abstract}
Modern LVLMs still struggle to achieve fine-grained document understanding, such as OCR/translation/caption for regions of interest to the user, tasks that require the context of the entire page, or even multiple pages. Accordingly, this paper proposes \textbf{Fox}, an effective pipeline, hybrid data, and tuning strategy, that catalyzes LVLMs to focus anywhere on single/multi-page documents. We introduce a novel task to boost the document understanding by making LVLMs focus attention on the document-level region, such as redefining full-page OCR as foreground focus. 
We employ multiple vision vocabularies to extract visual hybrid knowledge for interleaved document pages (\textit{e.g.}, a page containing a photo). Meanwhile, we render cross-vocabulary vision data as the catalyzer to achieve a full reaction of multiple visual vocabularies and in-document figure understanding. Further, without modifying the weights of multiple vision vocabularies, the above catalyzed fine-grained understanding capabilities can be efficiently tuned to multi-page documents, enabling the model to focus anywhere in both format-free and page-free manners. Besides, we build a benchmark including 9 fine-grained sub-tasks (\textit{e.g.}, region-level OCR/summary, color-guided OCR) to promote document analysis in the community. The experimental results verify the superiority of our model.

\end{abstract}

\section{Introduction}
\label{intro}
Recently, research on Large Vision-Language Models~\cite{GPT4,minigpt4,Flamingo} has been an attractive direction. These models not only easily handle some conventional vision tasks (\textit{e.g.}, Image Caption~\cite{coco_text}, OCR~\cite{OCRVQA}), but also demonstrate powerful reasoning capabilities like humans.

\begin{figure}[t]
\centering
\includegraphics[width=1.0\textwidth]{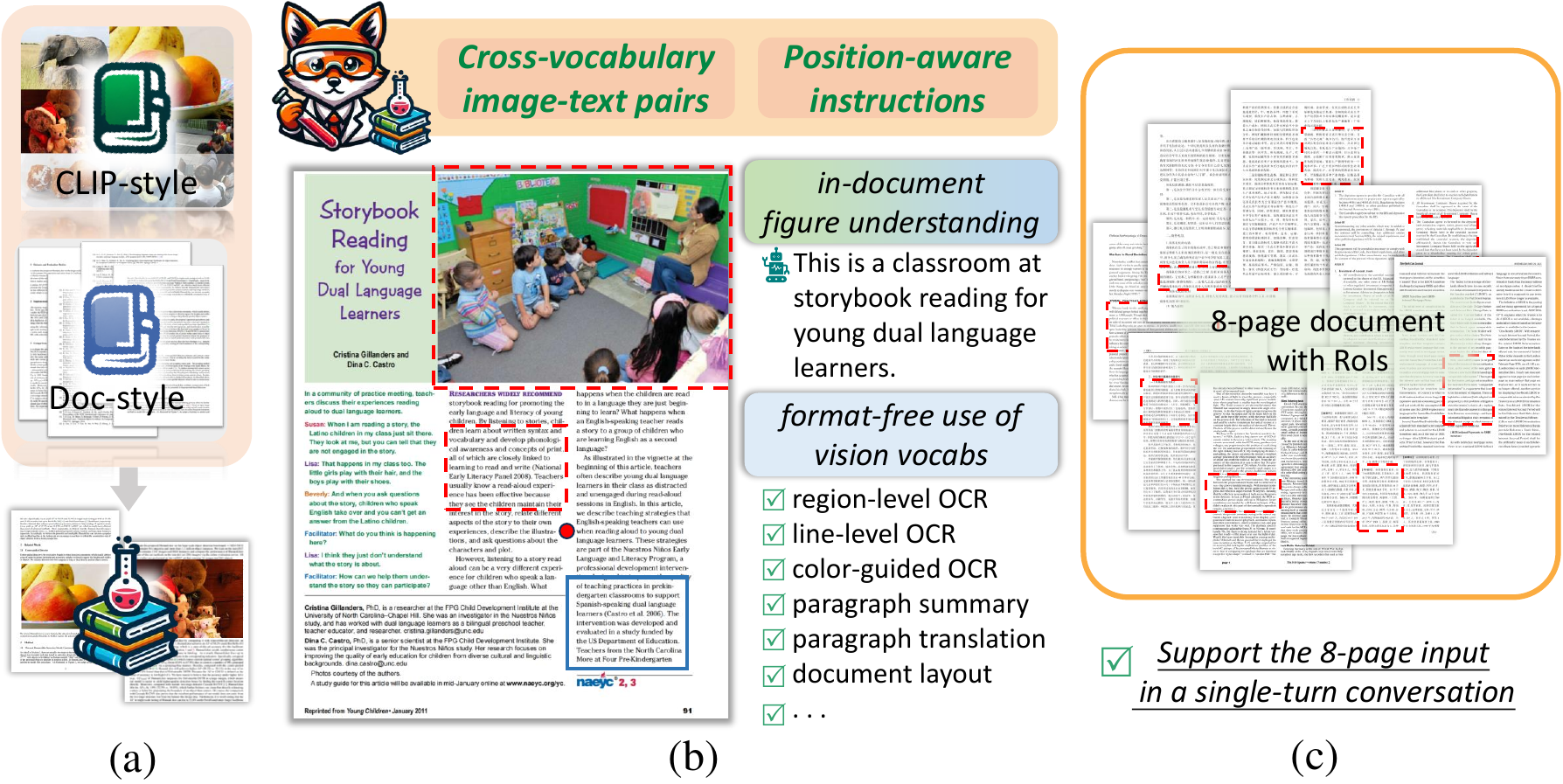}
\caption{(a) Multiple vision vocabularies are catalyzed using synthetic cross-vocabulary data to handle interleaved pages. (b) Fox achieves fine-grained document-level understanding by focusing anywhere, such as region-level OCR/translation and in-page figure caption. (c) Fox impressively supports the entire 8-page input and can focus on multiple cross-page RoIs in a single-turn conversation.}
\label{fig:intro}
\end{figure}

The LVLMs mostly give responses by leveraging large language models~\cite{OPT,vicuna,T5} to follow language instructions while referring to the vision vocabulary to understand the input image. 
Some researchers attempt to adopt LVLMs to advance the understanding of large-resolution (\textit{e.g.}, 833$\times$1132) document pages.  
For example, UReader~\cite{ye2023ureader} crops the input image into smaller patches to align with a CLIP-style vision vocabulary of input size 224$\times$224. Later, TextMonkey~\cite{liu2024textmonkey} divides the input image into 448$\times$448 patches and uses Openclip’s ViT-bigG~\cite{openclip_ilharco_2024_10469088} along with a resampling strategy to retain useful image tokens.
LLaVA-NeXT~\cite{liu2024llavanext} adopts CLIP-ViT-L-336px to perform visual perception and splits the input image into smaller patches to encode independently. InternVL-V1.5~\cite{chen2024far_intervl1.5} proposes a stronger vision vocabulary InternViT-6B with the input size of 448$\times$448. Similarly, to capture more details of the input image, InternVL-V1.5~\cite{chen2024far_intervl1.5} dynamically divides the input image into 1 to 12 tiles. Different from the methods above, without cropping patches, Vary~\cite{wei2023vary} writes an extra SAM-style~\cite{SAM} vision vocabulary specific to document and chart data, running in parallel with the CLIP branch. Vary can directly encode 1024$\times$1024 page into 256 image tokens with a high compression ratio.

The patch-based models~\cite{ye2023ureader,liu2024textmonkey,liu2024llavanext,chen2024far_intervl1.5} mostly employ CLIP-style vision vocabulary with small resolution, so a large-scale document needs to be decomposed into many patches/tiles. A patch/tile is independently encoded to 256 image tokens, and InternVL-V1.5~\cite{chen2024far_intervl1.5} even produces 3,328 image tokens during training. However, numerous image tokens are difficult to extend to multi-page documents for contextual understanding. More importantly, there may still be dense characters on these cropped patches, but CLIP-style vision vocabulary compresses limited sparse information of small input images via global contrastive learning, preventing these models from losslessly recovering the content of the original document (\text{e.g.}, full-page OCR).
Although Vary~\cite{wei2023vary} enjoys a high compression ratio and avoids cropping patches by directly encoding the document page, the lack of full collaboration across multiple vision vocabularies limits the performance. For example, given an input document page, Vary~\cite{wei2023vary} tends to only activate the SAM-style ViT branch due to the specific-vocabulary visual bias. In addition, the above models are sensitive to document format (\textit{e.g.}, multi-column) and do not support fine-grained user interaction on specific regions on documents.

Another key point for the document understanding is how to carry out fine-grained interaction, such as OCR/summarizing/captioning a region of interest. Actually, LVLMs with human-like referential dialogue capability for natural scenes have been investigated, such as Shikra~\cite{chen2023shikra} and ChatSpot~\cite{zhao2023chatspot}. They introduce referring spatial coordinates to refer to the special region of the input natural image, lifting the user experience and leading to more precise conversations. But these models can not handle the document images due to vision vocabulary CLIP-ViT~\cite{CLIP_radford2021learning} which is specific to natural scenes and has low input resolution. Besides, CLIP-style pre-training method based on Laion-COCO~\cite{schuhmann2021laion} (image-phrase pairs) only weakly write sparse visual knowledge, leading to a gap in understanding the dense document. Thus, we may ask: \textit{Can we devise an effective and efficient pipeline for LVLMs to achieve the fine-grained multi-page document understanding?}

In this paper, we propose Fox, an effective pipeline, hybrid data, and tunning strategy, giving a pleasing answer to the above question. The proposed Fox efficiently catalyzes the LVLM's attention to anywhere on single/multi-page documents in a user-friendly manner. Our solution has three highlights: 1) \textit{Focusing anywhere:} We introduce a novel task that boosts document understanding by focusing on the region of interest via fine-grained position-aware prompts, \textit{i.e.}, click points, dragged bounding boxes, and drawn color boxes. Notably, the dense full-page OCR sub-task can be further optimized by being redefined as foreground focus. 2) \textit{Full reaction across multiple vision vocabularies:} To fully interpret hybrid visual knowledge on interleaved document pages, we synthesize cross-vocabulary vision data to activate multiple visual vocabularies simultaneously to break down the specific-vocabulary bias of visual content, catalyzing multiple vision vocabularies to a full reaction. 
3) \textit{Supporting multi-column format and multiple pages:} With the position-aware prompts, the pipeline of focusing anywhere can yield robust performance regardless of document format. Moreover, benefiting from the high compression ratio (one 1024$\times$1024 page to 256 image tokes), we demonstrate the Fox can be efficiently tuned to achieve the above fine-grained capabilities on multi-page documents without modifying parameters of vision vocabulary.

As a result of the focusing catalytic process, the proposed Fox can not only give specific-vocabulary responses (\textit{e.g.}, page foreground OCR, region/line-level OCR/translation) but also gain the noticeable ability to utilize the cross-vocabulary visual knowledge (\textit{e.g.}, color-guided OCR, in-document figure caption). Furthermore, for more impressive multi-page document features, Fox can give the OCR results of {$region_1$} on $page_1$ and {$region_n$} on $page_n$ by only one question. Note that tasks like this with reference to cross-page content are of great research significance. We encourage researchers to rethink the framework design for LVLM-based document understanding and not be limited to conventional single-page sparse QA tasks.
Our contributions can be summarized as follows:
\begin{itemize}
\item We introduce a series of novel tasks to boost document understanding by enabling LVLMs to focus on document-level regions of interest. We propose an effective and efficient solution named Fox to focus anywhere on single/multi-page documents.
\item To catalyze multiple vision vocabularies for figure-text interleaved documents, we provide methods for generating hybrid data containing cross-vocabulary visual elements.
\item Fox is robust to documents of various formats due to the flexible position-aware prompts. Without training vision vocabulary, our Fox can be easily tuned to multi-page documents and gain cross-page parsing capabilities. 
\item We build a fine-grained document benchmark, including 9 sub-tasks, such as dense page OCR, region-level OCR/translation/summary, color-guided OCR, multi-page OCR/VQA. Experimental results show that our Fox outperforms other LVLMs by a large margin.
\end{itemize}

\section{Related Works}
\subsection{Visual Document Understanding}
Visual document understanding is widely investigated in the research field of computer vision. Optical Character Recognition (OCR) is a basic task, which plays a key role in document  digitalization~\cite{smith2007overview,moysset2017full}. The layout analysis task~\cite{zhong2019publaynet} aims to detect various document elements and facilitate to understanding of spatial relationships between them. We believe that OCR is a good task to test whether LVLMs can compress documents losslessly. Besides, for translation and summary~\cite{vaswani2017attention,dong2019unified} tasks, the proposed Fox can directly give answers for document images via the multimodal framework.

\subsection{Large Language Models}
In recent times, the success of LLMs has ignited the fields of natural language processing (NLP) and artificial general intelligence (AGI). 
The LLMs are built with the popular transformer framework which is explored by earlier NLP research, \textit{e.g.}, BERT~\cite{Bert}, GPT-2~\cite{GPT-2}, T5~\cite{T5}, and so on. Afterward, it is discovered that when the model parameters are expanded to a certain size, the language model will be greatly boosted due to the so-called "emergent ability"~\cite{wei2022emergent}. Further, the "GPT time" comes with amazing dialogue robots optimized by Reinforcement Learning with Human Feedback~\cite{RLHF_christiano2017deep}, \textit{e.g.}, InstructGPT~\cite{InstructGPT} and ChatGPT~\cite{ChatGPT}. Following that, OPT~\cite{OPT}, LLaMA~\cite{llama}, and GLM~\cite{GLM} are accessible to the community to pursue the performance like the GPT family. Based on the open-source LLMs, for more specific requirements, some fine-tuned models have merged, such as Alphaca~\cite{alpaca} and Vicuna~\cite{vicuna}, which also play critical roles in later Large Vision-Language Models.

\subsection{Large Vision-Language Models}
For vision-centric tasks, Large Vision-Language Models (LVLMs)~\cite{llava,Flamingo,lu2024deepseek} have been developed by connecting the vision networks to LLMs. CLIP-ViT~\cite{CLIP_radford2021learning} is a mature pre-trained vision vocabulary widely used to inject visual modality into language models. To ensure that LLMs can understand the visual context, LLaVA~\cite{llava} places the linear layers to project visual tokens into text space. Later, beyond natural scenes, LVLMs for large-resolution documents have emerged. 
UReader~\cite{ye2023ureader} is developed based on the LVLM mPLUG-Owl~\cite{ye2023mplug}. UReader~\cite{ye2023ureader} devise a shape-adaptive approach to crop input images into 224$\times$224 patches and feed them into CLIP vision encoder. 
Following Qwen-VL~\cite{Qwen-VL}, TextMonkey~\cite{liu2024textmonkey} uses a more powerful vision vocabulary Openclip’s ViT-bigG~\cite{openclip_ilharco_2024_10469088} with 448$\times$448 input size to endoce each cropped patch. With the strategy of cropping patches,
LLaVA-NeXT~\cite{liu2024llavanext} adopts CLIP-ViT-L-336px to perform visual perception.
Similarly, to capture more details, InternVL-V1.5~\cite{chen2024far_intervl1.5} dynamically divides the input image into 1 to 12 tiles of 448$\times$448. In contrast, without cropping patches, Vary~\cite{wei2023vary} writes an extra SAM-style~\cite{SAM} 1024$\times$1024 vision vocabulary specific to document and chart data, running in parallel with the CLIP branch. 

Compared to the above models, we believe that document understanding should move towards more fine-grained (\textit{e.g.,} region-level OCR/translation) and multi-page tasks. Imagine how cool it would be if we could use the LVLM like a reading pen! 
In this paper, we introduce Fox which can achieve fine-grained features by focusing anywhere on multi-page documents.

\section{Methods}
\label{methods}

\begin{figure}[t]
\centering
\includegraphics[width=1.0\textwidth]{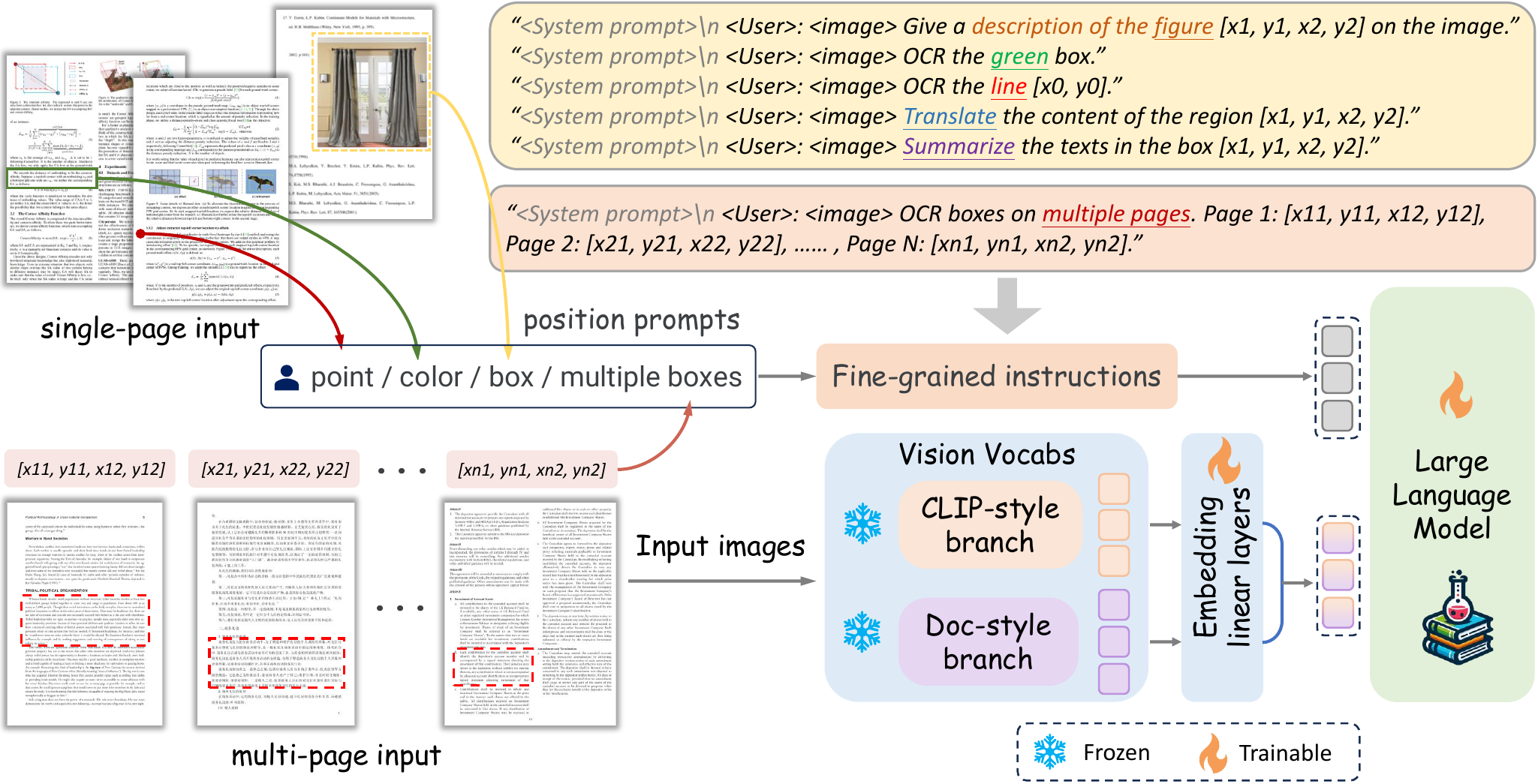}
\caption{Overall framework of the proposed Fox. All image tokens of multiple pages are unified into a sequence to achieve multi-page understanding. We devise position-aware prompts (point, color, and box) to make the model focus anywhere on single/multi-page documents.  We catalyze multiple vision vocabularies into a full reaction of hybrid visual knowledge for interleaved pages.}
\label{fig:architecture}
\end{figure}

In this section, we will elaborate on the details of the proposed Fox. First, we introduce the flexible pipeline which supports single/multi-page document understanding. Second, we provide the strategy to produce the data containing hybrid visual elements to activate multiple vocabularies concurrently. Last, we unify multi-task data with position-aware prompts to conduct the focusing process.

\subsection{Framework for Focusing Anywhere}
As illustrated in Figure~\ref{fig:architecture}, the architecture of the proposed Fox is built with two vision vocabularies, a large language model, and embedding linear layers. Specifically, to better handle figure-text interleaved large-resolution documents, there are two vision vocabularies, including natural content-aware CLIP-ViT~\cite{CLIP_radford2021learning} and artificial content-aware Vary-tiny~\cite{wei2023vary}. The overall framework is neat and provides more user-friendly fine-grained interactions, which can focus on the entire page and more specific regions of interest (RoI). Impressively, the proposed Fox also supports users to select RoIs on multiple pages at the same time, enabling cross-page contextual understanding.

Given a set of input document pages $I=\{p_i\}_{i=1}^N$, users can further indicate regions of interest $r_i$ on each page by clicking a point, dragging boxes, or drawing color boxes, and then give some language instructions $L^{instruct}$ about the questioning RoIs. $N$ is the number of input pages. The spatial coordinates or color information of $\{r_i\}_{i=1}^N$ is transformed into position-aware prompts and combined with $L^{instruct}$ to produce complete referential instructions. 
Meanwhile, two vision vocabularies will produce 256 image tokens $v^C_i \in \mathbb{R}^{256\times1024}$ and $v^S_i \in \mathbb{R}^{256\times1024}$ for each page $p_i$.  These image tokens $\{v^C_i\}_{i=1}^N$ and $\{v^S_i\}_{i=1}^N$ are sent into linear layers $W^C$ and $W^S$ to align with linguistic space. Then, the final image tokens $v_i \in \mathbb{R}^{256\times2048}$ can be obtained by concatenation. Note that $v_i$ is compressed into cross-vocabulary content, including dense characters and figures. Finally, with the projected image tokens and referential instructions, LLM will generate the response sequence $Q$ in an auto-regressive manner. The above process can be formulated as follows:
\begin{equation}
    \{v_i\}_{i=1}^N = \left[ W^C \circ \{v^C_i\}_{i=1}^N || W^S \circ \{v^S_i\}_{i=1}^N\right]
\end{equation}
\begin{equation}
    Q = \mathcal{LLM} \left( \{v_i\}_{i=1}^N, \left(L^{instruct}, \Psi \left(\{r_i\}_{i=1}^N \right)\right) \right)
\end{equation}
where $\left[\cdot || \cdot \right]$ is the concatenation operation. $\Psi(\cdot)$ denotes the normalization for spatial coordinates. Note that multi-page ($N$ pages) image tokens $\{v_i\}_{i=1}^N$ are unified into a sequence for cross-page contextual understanding.
With the causal masked sequence modeling, the training objective can be expressed as:
\begin{equation}
    \mathcal{L}_t=-E_{(Q, V)\sim D}\operatorname{log} P_{\theta} \left( q_m | q_{<m}, \{v_i\}_{i=1}^N \right)
\end{equation}
where $m$ denotes the current index of the target token and $D$ is the multi-page multi-grained dataset.

\subsection{Activating Multiple Vision Vocabularies by Cross-Vocabulary Hybrid Data} \label{sec:hybrid data} 
We hope to catalyze new capabilities more efficiently while freezing pre-trained multiple vision vocabularies.
Note that each vocabulary is written with visual knowledge of its specific-domain data. CLIP~\cite{CLIP_radford2021learning} using 400M~\cite{schuhmann2021laion} image-text pairs is geared toward perceiving natural images, and Vary-tiny~\cite{wei2023vary} using about 10M artificial data is good at page-level documents. There is a specific-vocabulary perceptual bias during querying vision vocabs due to the simple stacking of samples in two domains. 
Hence, we synthesize hybrid data containing the cross-vocabulary elements for the full reaction of multiple vision vocabularies. Instead of focusing too much on a specific vocabulary, richer cross-visual knowledge will be activated by breaking down the barriers of visual content.

\paragraph{Preparing PDF data.}  \label{para:prepare pdf}
We download enough open-source files in PDF format from e-books, CC-MAIN, and arXiv. Then, we parse the PDFs with some useful Python packages. We save each document page as a PNG-style image. Meanwhile, we locate each potential paragraph/line and record the corresponding bounding box and text content within it. 
\paragraph{Figure-text interleaved data.}
We choose the BLIP558k~\cite{llava}, Laion-COCO~\cite{schuhmann2021laion}, and RegionChat~\cite{zhao2023chatspot} datasets that contain descriptions for natural images, and we randomly sample the same number of document pages from the prepared PDF data. Clearly, before we render a natural image of $W^n\times H^n$ pixels into a document page with the resolution of $W^d \times H^d$ pixels, we should make the size of the natural image smaller than that of the document page. The scaling process can be formulated as follows:
\begin{small}
\begin{equation}
\label{eq1}
\left\{ \begin{aligned}
    W_{new}^n & = \operatorname{randint}\left(\left[\alpha \cdot W^d \right], \left[\beta \cdot W^d\right] \right), H_{new}^n = \left[W_{new}^n/W^n \cdot H^n \right], & \text{if} \  W^n/H^n > W^d/H^d \\
    H_{new}^n & = \operatorname{randint}\left(\left[\eta \cdot H^d \right], \left[\gamma \cdot H^d\right] \right), W_{new}^n = \left[H_{new}^n/H^n \cdot W^n \right], & \text{if} \ W^n/H^n \leq W^d/H^d\\
\end{aligned} \right.
\end{equation}
\end{small}
where $W_{new}^n$/$H_{new}^n$ denote the desired width/height of the scaled natural image.  $\left[\cdot\right]$ means the integral function. $\alpha$, $\beta$, $\eta$, and $\gamma$ are the hyperparameters that control the scaling ratio, and they are set to 0.3, 0.9, 0.4, and 0.9, respectively. 
Then, we randomly pick a suitable location $(x^n_1, y^n_1, x^n_2, y^n_2)$ on the page to place the scaled natural image. What's more, to make the interleaved data reasonable and delete the occluded text on this page, we calculate the intersection of union (IoU) between $(x^n_1, y^n_1, x^n_2, y^n_2)$ and the vanilla text boxes $\left\{ (x^d_{i,1}, y^d_{i,1}, x^d_{i,2}, y^d_{i,2}) \right\}_{i=1}^{N_d}$, and fill the text boxes overlapped by the natural image with the white color. $N_d$ is the number of text boxes on this document page. So, we can obtain cross-vocabulary image-text pairs for in-document figure caption. The text for each interleaved page includes the filtered optical characters and the description of the pasted natural image.

\paragraph{Color-text hybrid data.}
CLIP is written with the knowledge for recognizing colors, while the Vary-tiny is not. We produce color-text hybrid data to further activate multiple vocabularies, which is the key to enabling Fox to support the conversations for users' color-guided RoI.
We randomly select three text boxes and paint them directly on the document page in red, blue, and green colors. The proposed Fox is expected to directly give the OCR results in the area with the questioning color.

\subsection{Triggering Focusing Process via Fine-grained Instruction-following Tasks}
We devise fine-grained instructions based on several position-aware text prompts, such as points, boxes, and colors, to catalyze Fox to focus any fine-grained region on single/multi-page documents.

\paragraph{Fine-grained document understanding.}
We define several novel sub-tasks to drive the model to focus on fine-grained regions for flexible document-level understanding: 1) Foreground OCR. We redefine the page OCR task as the foreground focus to further boost the dense perception. The instruction can be ``\textit{Give the OCR results of the box $(x^f_{i,1}, y^f_{i,1}, x^f_{i,2}, y^f_{i,2})$}''. The foreground box can be obtained by some simple operations. 2) Region-level OCR. Based on the obtained text boxes, we transform the content of one page into multiple region-level OCRs via multi-turn conversations. An example can be ``\textit{Give the OCR results of the box $(x^d_{i,1}, y^d_{i,1}, x^d_{i,2}, y^d_{i,2})$}''. 3) Line-level OCR. We pick a point near the left side of each line as the position prompt. Then, we construct the line-level multi-turn conversations and an example can be like ``\textit{OCR the line $(x^d_{j}, y^d_{j})$}''.  4) Color-guided OCR. Using the color-text hybrid data in Section~\ref{sec:hybrid data}, we define the corresponding cross-vocabulary task by some color-guided questions, such as ``\textit{OCR red box}'' and ``\textit{OCR blue box}''. 5) Region-level translation and summary. We filter and retain the boxes with text lengths over 400 on each page. Then, we employ GPT-3.5~\cite{ChatGPT} to generate the translation and summary for each long in-box text as the corresponding annotations. The instruction can be ``\textit{Translate/Summarize the content of the box $(x^d_{i,1}, y^d_{i,1}, x^d_{i,2}, y^d_{i,2})$}''.
6) Document layout: We convert the 330K high-quality annotations of PubLayNet~\cite{zhong2019publaynet} to the unified conversation format. Further, we sample 1M extra PDF pages and use PaddleOCRv2~\cite{paddleocrv2_du2021pp} tools to generate pseudo layout annotations.

\paragraph{In-document figure understanding.}
Based on the synthetic interleaved data, we organize the cross-vocabulary image-text pairs into two sub-tasks: 1) In-document figure caption. As a result of the added position-aware prompts, an example language instruction is as follows: ``\textit{Give a brief description for the region $(x^n_1, y^n_1, x^n_2, y^n_2)$ of the image}''. The box denotes the boundary of the figure.
2) In-document in-figure chat. The RegionChat~\cite{zhao2023chatspot} dataset is built for referential dialogue on natural images. After rendering it on PDF pages, with spatial coordinates of the referring region, we can ask the proposed Fox the following question: ``\textit{What can you see in this region? $(x^n_1, y^n_1, x^n_2, y^n_2)$}''. At a more fine-grained level, the RoI can be the box within the figure on the document page.

\paragraph{Extension for multi-page documents.}
The proposed Fox can be easily tuned to focus on multiple regions of multi-page documents using simple instructions. As a forerunner, we define two basic yet interesting multi-page sub-tasks and give position-aware instruction examples. 1) Multi-page region-level OCR: ``\textit{OCR boxes on multiple pages. Page 1: $(x^1_1, y^1_1, x^1_2, y^1_2)$, Page 2: $(x^2_1, y^2_1, x^2_2, y^2_2)$, $\dots$ Page N: $(x^N_1, y^N_1, x^N_2, y^N_2)$}''. 2) Cross-page VQA: ``\textit{Which page's box contains more characters? Page 1: $(x^1_1, y^1_1, x^1_2, y^1_2)$, Page 2: $(x^2_1, y^2_1, x^2_2, y^2_2)$, $\dots$ Page N: $(x^N_1, y^N_1, x^N_2, y^N_2)$}''.

It is worth noting that all the above methods are independent of document format. The PDF data with any format or layout, such as single-column, double-column, interleaved, \textit{etc.}, can be parsed to extract positional prompts and formulated into the corresponding conversations.
With the fine-grained position-aware instructions, the vision query pipeline enjoys high human-AI interactivity and is robust to different formats (multi-column) and multi-page documents.

\subsection{Catalyzing Fox by Multi-page and Multi-grained Data Engine}
The data engine is a key part of the proposed Fox. To ensure the performance on multiple tasks, We carefully control the quantity and ratio of training data, and more details are reported in Table~\ref{tab:data}.

\paragraph{Pre-training data.}
In the pre-training stage, we formulate a large number of multimodal task-driven data. Specifically, for hybrid images of in-document caption and chat sub-tasks, we render the BLIP558K~\cite{llava} data, 1M natural images sampled in Laion-COCO~\cite{schuhmann2021laion} and RegionChat100K~\cite{zhao2023chatspot} data into an equal amount of document pages sampled in prepared PDF data. For fine-grained optical character understanding, we formulate 6 types of 4.6M document image-text pairs, containing box/line/color position-aware prompts and OCR/translation/summary interactive task forms. Further, we generate 800K multi-page data, including multi-page multi-region OCR and cross-page QA.
In addition, to maintain the general conversational capabilities of our model, we sample 1M natural data from Laion-COCO~\cite{schuhmann2021laion} and NLP dialogue data from Alpaca~\cite{alpaca}, Baize~\cite{xu2023baize} and ShareGPT.

\paragraph{SFT data.}
In the supervised fine-tuning stage, To make the conversation experience more comfortable, we sample 10K image-text pairs for each data type of the above pre-training data, and adopt GPT3.5~\cite{ChatGPT} to rewrite prompts ten times more diversified. Besides, LLaVA80K~\cite{llava} is also added to further tune our model to generate pleasing answers. 

\begin{table*}[h]
\footnotesize
\caption{Details of multi-page and multi-grained data. In the pre-training phase, we use customized region-level document-text pairs to bring out focusing capabilities. 
}
\begin{center}
\setlength{\tabcolsep}{2pt}
{
\begin{tabular}{c|c|c|c|c|c}
\toprule[.9pt]
{\textbf{Task}} & \bf{Region-level Dataset} & \bf{Sample} & {\textbf{Task}} & \bf{Page-level Dataset} & \bf{Sample}  \\  
\midrule
\multirow{2}{*}{In-document Cap.} & PDF$\times$BLIP558K~\cite{llava} & 558K & \multirow{2}{*}{Layout} & PubLayNet~\cite{zhong2019publaynet} & 33K \\
& PDF$\times$ Laion-COCO~\cite{schuhmann2021laion} & 1M & & Annots. by PaddleOCRv2~\cite{paddleocrv2_du2021pp} & 1M \\
\midrule
{In-document Chat} & PDF$\times$ RegionChat~\cite{zhao2023chatspot} & 22K & Cap.& Laion-COCO~\cite{schuhmann2021laion} & 500K \\
\midrule
\multirow{6}{*}{Doc. Understanding} & foreground OCR & 1M & \multirow{3}{*}{NLP} & Alpaca~\cite{alpaca} & 52K \\
& region-level OCR & 1M & & Baize~\cite{xu2023baize} & 112K\\
& line-level OCR & 600K & & ShareGPT & 125K\\
& color-guided OCR & 1M & --------- & ------------------------------------ & ------------- \\
& region-level translation & 500K & \multirow{2}{*}{PDF} & Page OCR & 1M\\
& region-level summary & 500K & & Page Markdown & 1M \\
\midrule
\multirow{2}{*}{Multi-page Doc.} &  multi-region OCR & 400K & - & - & -\\
& cross-page VQA & 400K & - & - & -\\

\bottomrule[.9pt]
\end{tabular}

}
\end{center}
\label{tab:data}
\end{table*}

\paragraph{Input and Conversation Format}
For each input image, we resize it with a fixed resolution 1024$\times$1024 before feeding it into the SAM-style~\cite{SAM} ViT branch and we perform a resize operation to obtain a new image of 224$\times$224 as the input of the CLIP vision network. We choose Qwen-1.8B~\cite{qwen} with rich linguistic vocabulary as our language model. Following the LLaVA-MPT~\cite{llava,team2023introducing} dialogue style, the input conversation format can be summarized as follows: <|im\_start|>user: <img>"<image>"</img> "\textit{human question [position-aware prompts]}"<|im\_end|> <|im\_start|>assistant: "\textit{AI responses}" <|im\_end|>.

\section{Experiments}
\subsection{Implementation Details}
During the multi-task pre-training and SFT phase, the multiple vision vocabularies (CLIP and SAM-style ViT) are frozen and only the parameters of the embedding linear layers and language model are optimized. We train our model using the optimizer AdamW~\cite{AdamW} and a cosine annealing scheduler~\cite{loshchilov2016sgdr}. The learning rate is set to 1e-4 in pretraining and then to 2e-5 in SFT. In both stages, we use 48 A800 GPUs with a per device batch of 4 and the data epoch is set to 1.

\subsection{Multi-grained Benchmark and Metrics}
To advance fine-grained document understanding, we build a bilingual benchmark including 9 sub-tasks. We collect 112 English pages and 100 Chinese pages, including single/multi-column formats. The number of words per page exceeds 1,000. These images are used to evaluate page OCR, line-level OCR, color-guided OCR, region-level OCR/translation/summary, multi-page multi-region OCR, and cross-page VQA. Besides, to monitor the performance of interleaved data, we render 200 natural images sampled from Laion-COCO~\cite{schuhmann2021laion} onto 200 PDF pages to evaluate the document-level in-figure caption task. The comprehensive evaluation metrics contain normalized edit distance, F1-score, BLEU~\cite{papineni2002bleu}, METEOR~\cite{banerjee2005meteor}, ROUGE~\cite{lin2004rouge}, and \textit{etc}.

\begin{table}[h]
\footnotesize
\caption{Dense English text recognition on the single document page.}
\label{tab:en_page_ocr}
\centering
\setlength{\tabcolsep}{3pt}
{
\begin{tabular}{lccccccc}
\toprule[.9pt]
{\textbf{Method}} & Params & Edit Distance  $\downarrow$ & {F1-score} $\uparrow$  &  Precision $\uparrow$ & Recall $\uparrow$ & BLEU $\uparrow$ & METEOR $\uparrow$  \\  
\midrule
LLaVA-NeXT~\cite{liu2024llavanext} & 34B & 0.430 & 0.647 & 0.573 & 0.881 & 0.478 & 0.582 \\
{InternVL-ChatV1.5~\cite{chen2024far_intervl1.5}} & 26B & 0.393 & 0.751 & 0.698 & 0.917  & 0.568 & 0.663\\ 
{Nougat~\cite{blecher2023nougat}} & 250M & 0.255 & 0.745 & 0.720 & 0.809 & 0.665 & 0.761 \\ 
Vary~\cite{wei2023vary} & 7B & 0.092 & 0.918 & 0.906 & 0.956 & 0.885 & 0.926  \\
{Vary-toy~\cite{wei2024small_varytoy}} & 1.8B & 0.082 & 0.924 & 0.919 & 0.938 & 0.889 & 0.929 \\  
Qwen-VL-Plus~\cite{Qwen-VL} & >100B & 0.096 & 0.931 & 0.921 & 0.950 & 0.893 & 0.936 \\
{Qwen-VL-Max~\cite{Qwen-VL}} & >100B & 0.057 & \bf{0.964} & 0.955 & \bf{0.977} & \bf{0.942} & \bf{0.971} \\ 
\rowcolor{gray!10}
{Fox (foreground focus)} & \bf{1.8B} &  \bf{0.046} &  0.952  &  \bf{0.957} &  0.948  & 0.930  &  0.954 \\ 
\bottomrule[.9pt]
\end{tabular}
}
\end{table}

\begin{table}[h]
\footnotesize
\caption{Dense Chinese text recognition on the single document page.}
\label{tab:cn_page_ocr}
\centering
\setlength{\tabcolsep}{2pt}
{
\begin{tabular}{lccccccc}
\toprule[.9pt]
{\textbf{Method}} & Params & Edit Distance  $\downarrow$ & {F1-score} $\uparrow$  &  Precision $\uparrow$ & Recall $\uparrow$ & BLEU $\uparrow$ & METEOR $\uparrow$  \\  
\midrule
{InternVL-ChatV1.5~\cite{chen2024far_intervl1.5}} & 26B & 0.265 & 0.816 & 0.784 & 0.866 & 0.622 & 0.717   \\ 
{Vary-toy~\cite{wei2024small_varytoy}} & 1.8B & 0.142 & 0.914 & 0.928 & 0.907 & 0.718 & 0.832 \\  
Qwen-VL-Plus~\cite{Qwen-VL} & >100B & 0.121 & 0.895 & 0.903 & 0.890 & 0.684 & 0.828\\
Vary~\cite{wei2023vary} & 7B & 0.113 & 0.952 & 0.961 & 0.944 & 0.754 & 0.873 \\
{Qwen-VL-Max~\cite{Qwen-VL}} & >100B & 0.091 & 0.931  &  0.917  &  0.946 & 0.756 & 0.885 \\ 
\rowcolor{gray!10}
{Fox (foreground focus)} & \bf{1.8B} & \bf{0.061} &  \bf{0.954} & \bf{0.964}  & \bf{0.946} & \bf{0.842} & \bf{0.908} \\ 
\bottomrule[.9pt]
\end{tabular}
}
\end{table}

\begin{table}[!h]
\footnotesize
\caption{The performance of fine-grained text recognition on the single page. The focusing regions are given by drawn color boxes, coordinates of region boundary, and line points, respectively.}
\label{tab:boxline}
\centering
\setlength{\tabcolsep}{9pt}
{
\begin{tabular}{clcccccc}
\toprule[.9pt]
\multirow{2}{*}{\textbf{Method}}& \multirow{2}{*}{\textbf{Forms}} & \multicolumn{3}{c}{\textbf{English}} & \multicolumn{3}{c}{\textbf{Chinese}}  \\ 
\cmidrule(rl){3-5} \cmidrule(rl){6-8} & & color & region & line  & color & region & line \\  \midrule
\multirow{6}{*}{\thead{Fox\\(region focus)}}  
& Edit Distance $\downarrow$  & 0.064 & 0.059 & 0.116 & 0.114 & 0.042 & 0.084 \\ 
& F1-score $\uparrow$ & 0.940 & 0.957 & 0.879 & 0.884 & 0.955 & 0.918 \\
& Precision $\uparrow$  &  0.942 & 0.962 & 0.879 & 0.902 & 0.966 & 0.931\\ 
& Recall $\uparrow$ &   0.942 & 0.955 & 0.883 & 0.873 & 0.947 & 0.909  \\ 
& BLEU $\uparrow$  & 0.868 & 0.914 & 0.845 &  0.778 & 0.885 & 0.825\\
& METEOR $\uparrow$  & 0.938 & 0.955 & 0.878 & 0.848 & 0.934 & 0.886 \\
\bottomrule[.9pt]
\end{tabular}
}
\end{table}

\subsection{Evaluation Results}
\paragraph{Foreground focus for dense text recognition on a single page.}
For the dense text recognition on the entire page, we directly input the normalized box $\left[2, 2, 998, 998\right]$ as the foreground prompts. As shown in Table~\ref{tab:en_page_ocr} and ~\ref{tab:cn_page_ocr}, Fox showcases strong English and Chinese dense OCR ability by almost lossless compression for the document page. Specifically, Fox achieves the best edit distance of 0.046 and 0.061 in English and Chinese, respectively. Compared to Vary-toy using the image-level prompts, the proposed Fox lifts the English F1-score by 2.8\% by redefining the task as foreground focus. Note that the performance of LLaVA-NeXT and InternVL-ChatV1.5 which use the CLIP-style vocabulary is bottle-necked, indicating that the dense texts of each patch are not completely encoded.

\paragraph{Region focusing performance of in-document fine-grained tasks.}
As shown in Table~\ref{tab:boxline}, Fox can yield excellent OCR results on various metrics under several color-guided/region-level/line-level settings, indicating that our model can accurately recognize the content in these randomly sampled RoIs.
In Table~\ref{tab:translation_summary}, for the region-level translation, Fox yields an acceptable METEOR of 0.366 due to the smaller language model of 1.8B parameters. In addition, we evaluate our model on the fine-grained summary task and obtain a decent ROUGE-L-F score of 0.282. It is worth mentioning that this kind of usage similar to a reading pen is exactly what users need more.

\begin{table}[ht]
\footnotesize
\caption{The performance of in-document fine-grained understanding tasks. The fine-grained translation/summary/caption tasks are targeted at interpreting in-document text/figure regions.}
\label{tab:translation_summary}
\centering
\setlength{\tabcolsep}{5pt}
{
\begin{tabular}{ccccccc}
\toprule[.9pt]
 \multicolumn{2}{c}{\textbf{Fine-grained Translation}} & \multicolumn{3}{c}{\textbf{Fine-grained Summary}} & \multicolumn{2}{c}{\textbf{Fine-grained Caption}} \\ 
\cmidrule(rl){1-2} \cmidrule(rl){3-5} \cmidrule(rl){6-7} BLEU   & METEOR  & ROUGE-L R & ROUGE-L P & ROUGE-L F & METEOR & ROUGE-L F \\ 
\midrule
0.138 & 0.366 & 0.261 & 0.316 & 0.282 & 0.359 & 0.396\\ 

\bottomrule[.9pt]
\end{tabular}
}
\end{table}

\begin{table}[!h]
\footnotesize
\caption{The performance of fine-grained tasks on the multi-page (8 pages) documents.}
\label{tab:multipage}
\centering
\setlength{\tabcolsep}{6pt}
{
\begin{tabular}{lccccc}
\toprule[.9pt]
\multirow{2}{*}{\textbf{Method}} & \multicolumn{4}{c}{\textbf{Multi-page (8 pages) multi-region OCR}} & {\textbf{Cross-page (8 pages) VQA}}  \\ 
\cmidrule(rl){2-5} \cmidrule(rl){6-6} & Edit Distance $\downarrow$ & F1-score  $\uparrow$ & BLEU  $\uparrow$ & METEOR  $\uparrow$  & Accuracy  $\uparrow$ \\ 
\midrule
{Fox (Ours)}  & 0.084  & 0.946 & 0.836 & 0.805 & 0.827 \\ 
\bottomrule[.9pt]
\end{tabular}
}
\end{table}

\paragraph{Cross-vocabulary focusing tasks on interleaved pages.}
The color-guided task requires cross-vocabulary visual knowledge, \textit{i.e.}, CLIP for recognizing colors and Vary-tiny for capturing texts. Table~\ref{tab:boxline} shows that the decent results (0.940 and 0.884 on English and Chinese F1-score) meet our expectations due to the collaboration across multiple vision vocabularies. 
For the in-document figure caption task, we render natural images onto document pages and ask our model ``\textit{What is this in the box $<box>$?}'', where $<box>$ is the boundary of the natural image that is pasted into the document page. As shown in Table~\ref{tab:translation_summary}, when handling interleaved data, Fox reaches the METEOR of 0.359 and ROUGE-L-F of 0.396 due to the full reaction of activating multiple vocabularies. 

\paragraph{Exploration for focusing on multiple pages.}
To verify the focusing capability of Fox on multi-page documents, we report two relevant results in Table~\ref{tab:multipage}. For the multi-page OCR task, we ask the model to output the OCR results of 8 boxes on 8 complex pages (in mixed English/Chinese and mixed single/multi-column formats) in a single-turn conversation. Our Fox still performs an amazing F1-score of 0.946 and achieves true focus anywhere by parsing the entire 8-page document simultaneously. For the cross-page visual question-answering task which requires the model to answer which box has the largest number of characters in multiple cross-page boxes, Fox yields a high accuracy of 0.827, demonstrating that it is easier to perform VQA reasoning based on successfully perceiving dense text of multiple pages.

\begin{figure}[!h]
\centering
\includegraphics[width=1.0\textwidth]{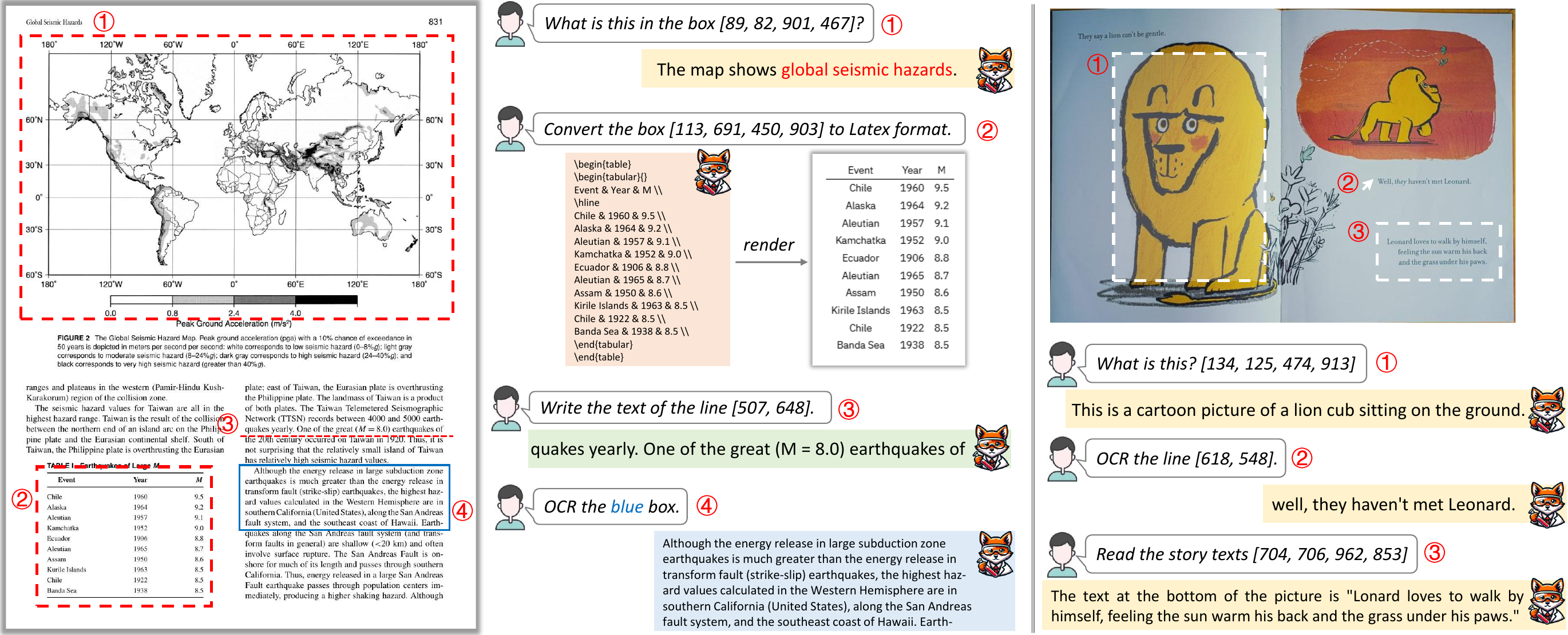}
\caption{Visualization results. Fox can focus anywhere by supporting fine-grained features, such as in-document figure caption, color-guided OCR, VQA in the cartoon book, and \textit{etc}. }
\label{fig:vis}
\end{figure}

\paragraph{Visualization.}
Figure~\ref{fig:vis} shows our Fox can perform impressive features with high human-AI interactivity. For the figure on the academic page, Fox gives the response ``global seismic hazards'' which is relevant to the content of the document. Fox can also give precise OCR results by dense text perception. For the cartoon book, Fox can recognize the interesting ``lion'' and can read the story texts for users. This indicates that our Fox enjoys fine-grained focusing capabilities in various scenarios.

\section{Conclusion and Limitations}
\label{discussion}
This paper proposes a user-friendly LVLM named Fox, which enjoys amazing fine-grained capabilities of focusing anywhere on single/multi-page documents. Further, after catalyzing the multiple vision vocabularies into a full reaction, Fox gains impressive cross-vocabulary features on figure-text interleaved pages. To advance the fine-grained document understanding, we provide a benchmark containing comprehensive sub-tasks. Our Fox can achieve promising scores in these experiments, making a successful step to high human-AI interactivity on dense-content documents. We believe that the proposed method has considerable room for improvement (\textit{e.g.}, the low-resolution CLIP), and we encourage more researchers to focus on more reasonable multi-page document-level tasks.


\section{Appendix}
We show more amazing output results of our model Fox. All testing images are from the Internet.
\begin{figure}[!h]
\centering
\includegraphics[width=1.0\textwidth]{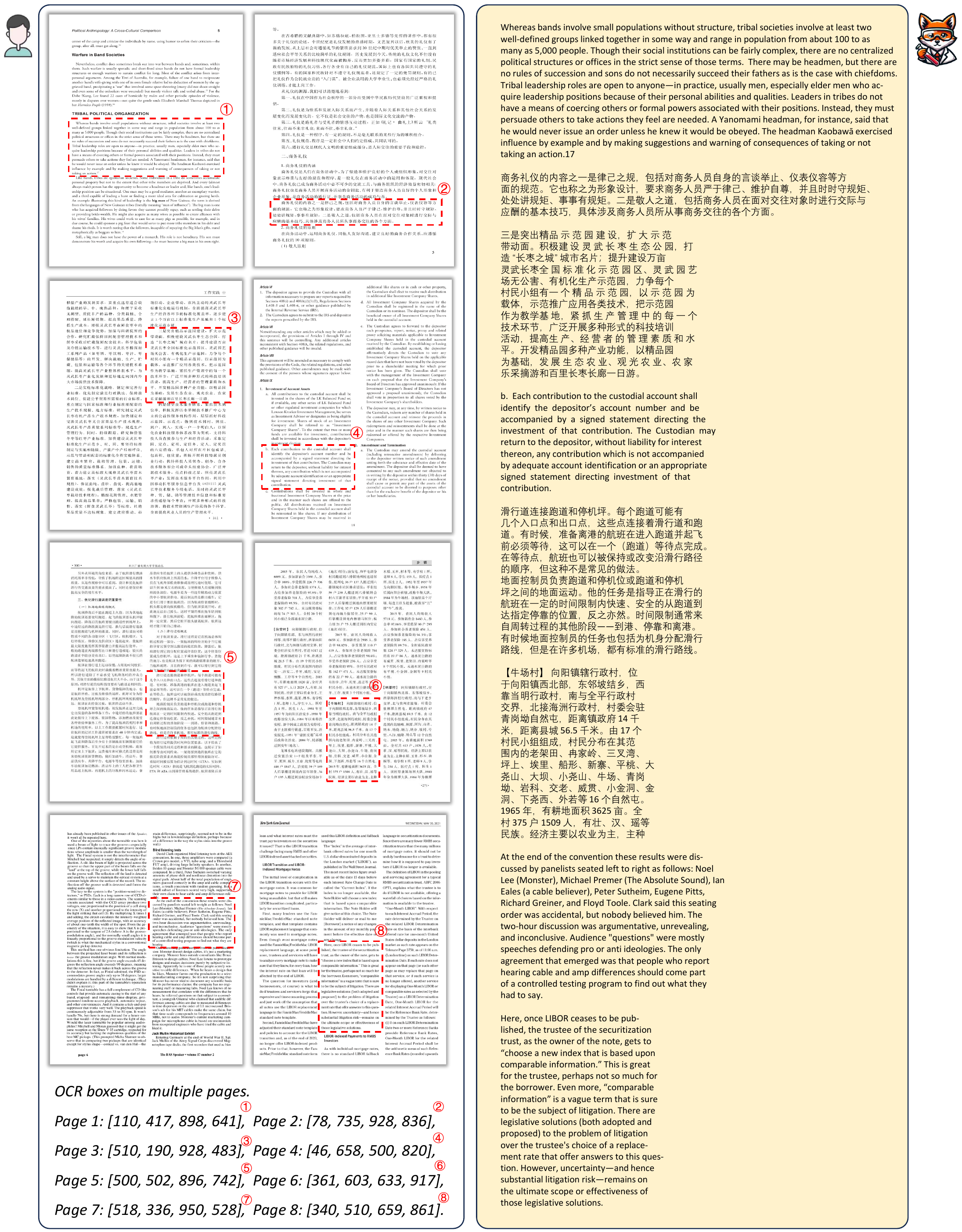}
\caption{Fox can give precise responses when focusing on the 8-page document. These pages contain bilingual content, have well over a thousand characters per page, and have a variety of single and multi-column layouts. This extreme case demonstrates powerful focusing capabilities.}
\label{fig:append1}
\end{figure}

\newpage

\begin{figure}[!h]
\centering
\includegraphics[width=1.0\textwidth]{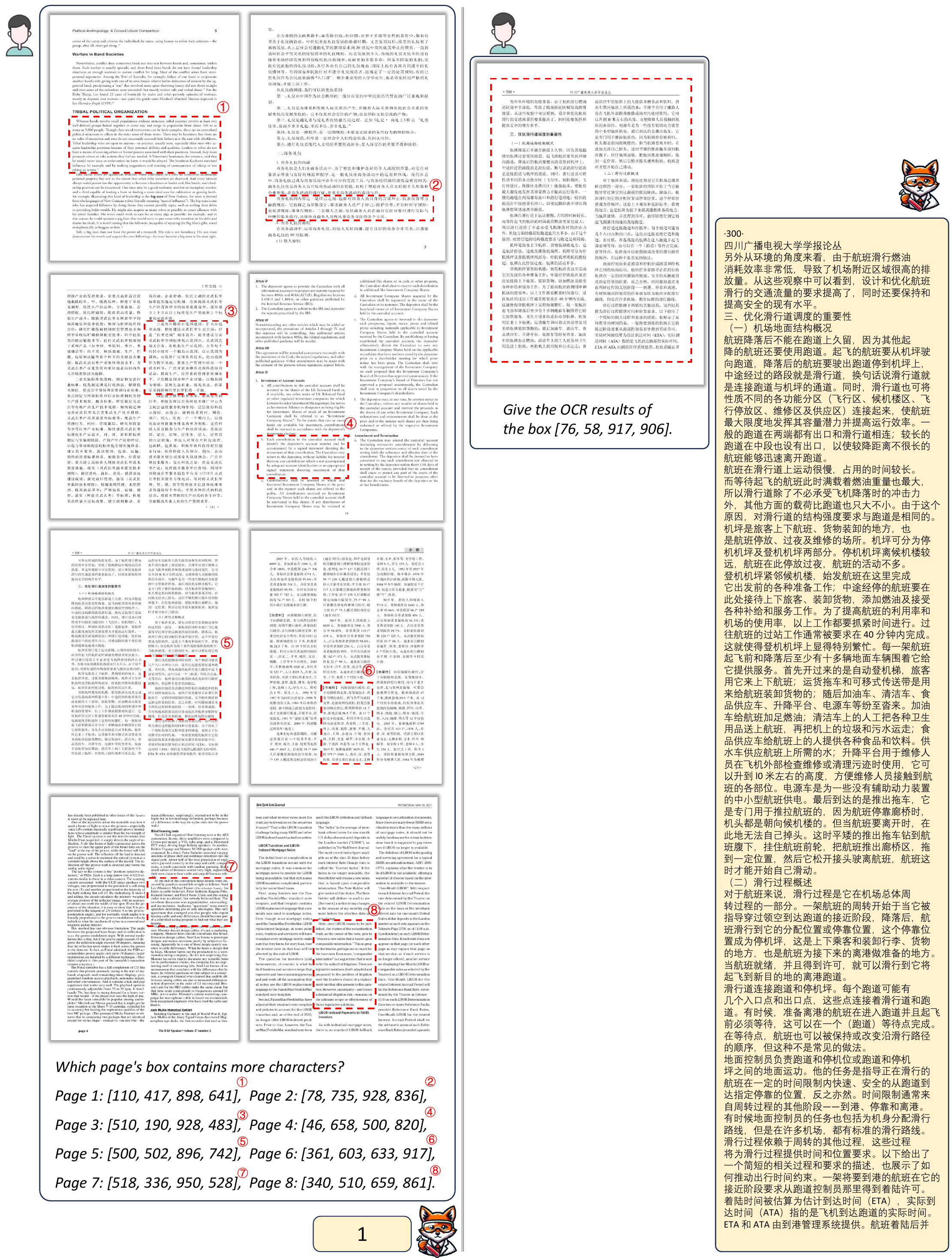}
\caption{The left case shows Fox can handle the cross-page VQA task on the multi-page (8 pages as an example) document. The right case shows Fox can perform the dense Chinese text recognition by foreground focus and obtain precise results.}
\label{fig:append2}
\end{figure}

\newpage

\begin{figure}[!h]
\centering
\includegraphics[width=1.0\textwidth]{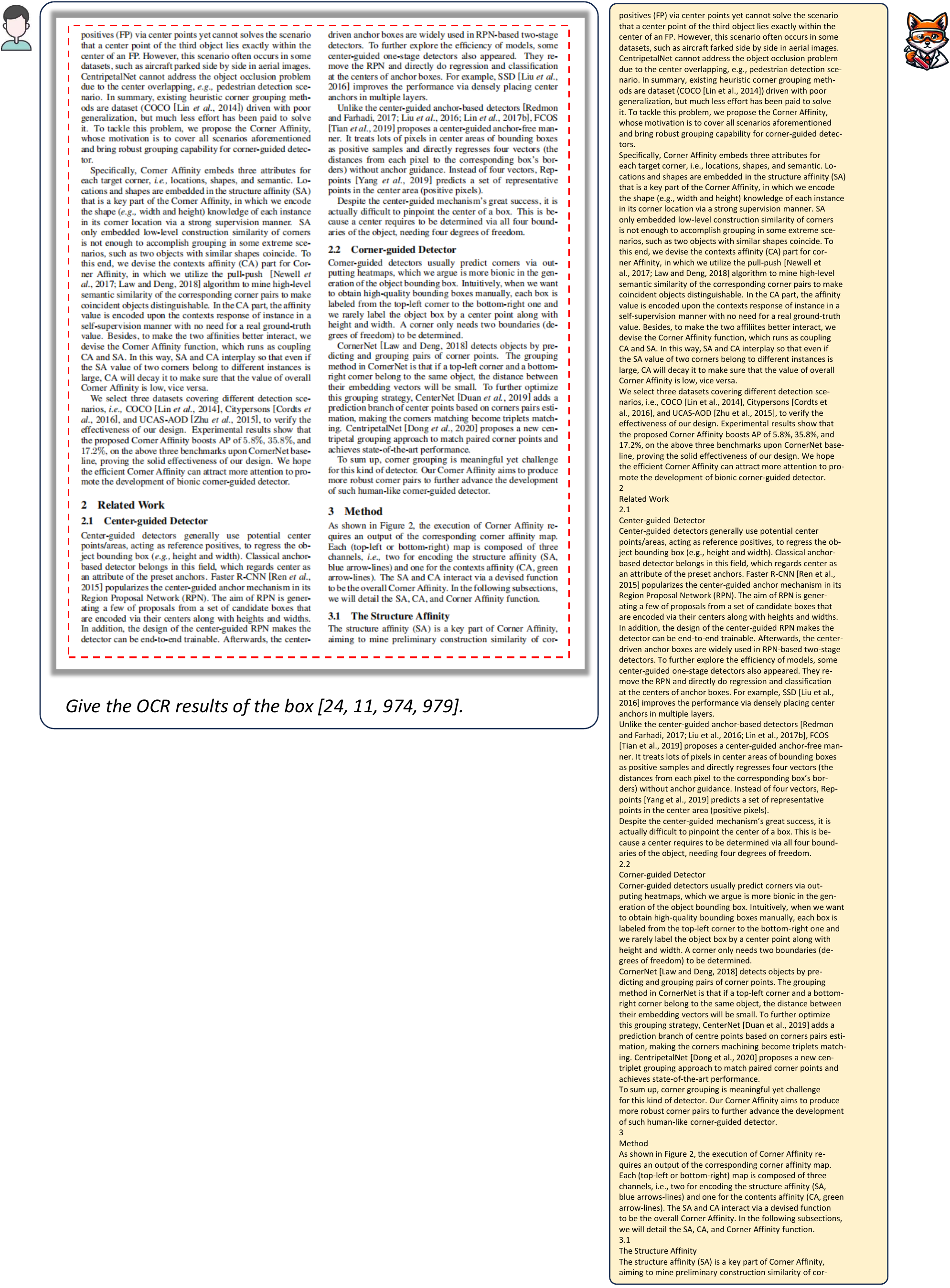}
\caption{The proposed Fox easily performs dense English text recognition by foreground focus.}
\label{fig:append3}
\end{figure}

\newpage

\begin{figure}[!h]
\centering
\includegraphics[width=1.0\textwidth]{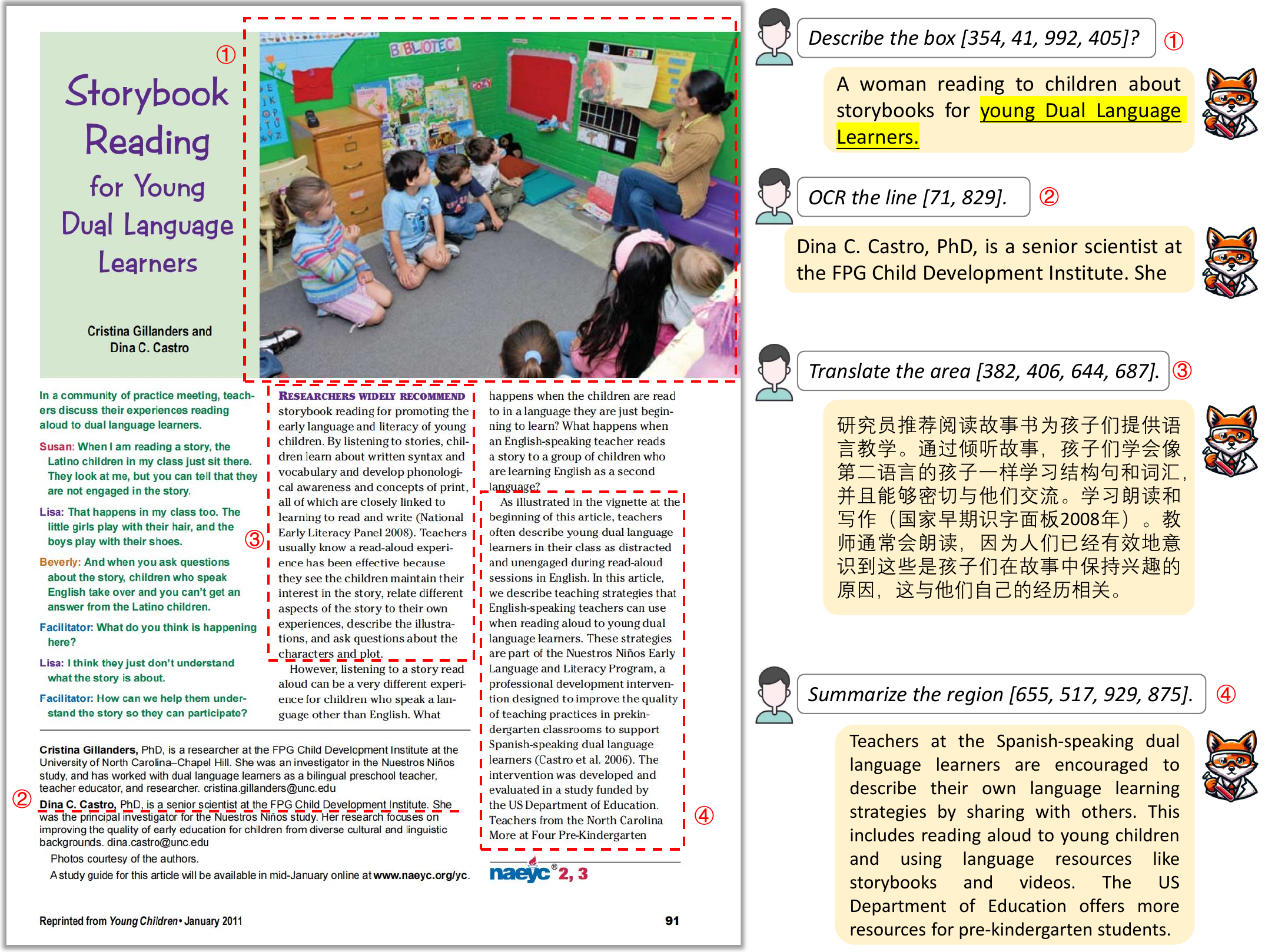}
\caption{Fox can achieve text-associative in-page figure caption and fine-grained document understanding. Fox enjoys high flexibility and robustness when performing fine-grained region-level translation/summary/OCR tasks in multi-column documents.}
\label{fig:append4}
\end{figure}

\begin{figure}[!h]
\centering
\includegraphics[width=1.0\textwidth]{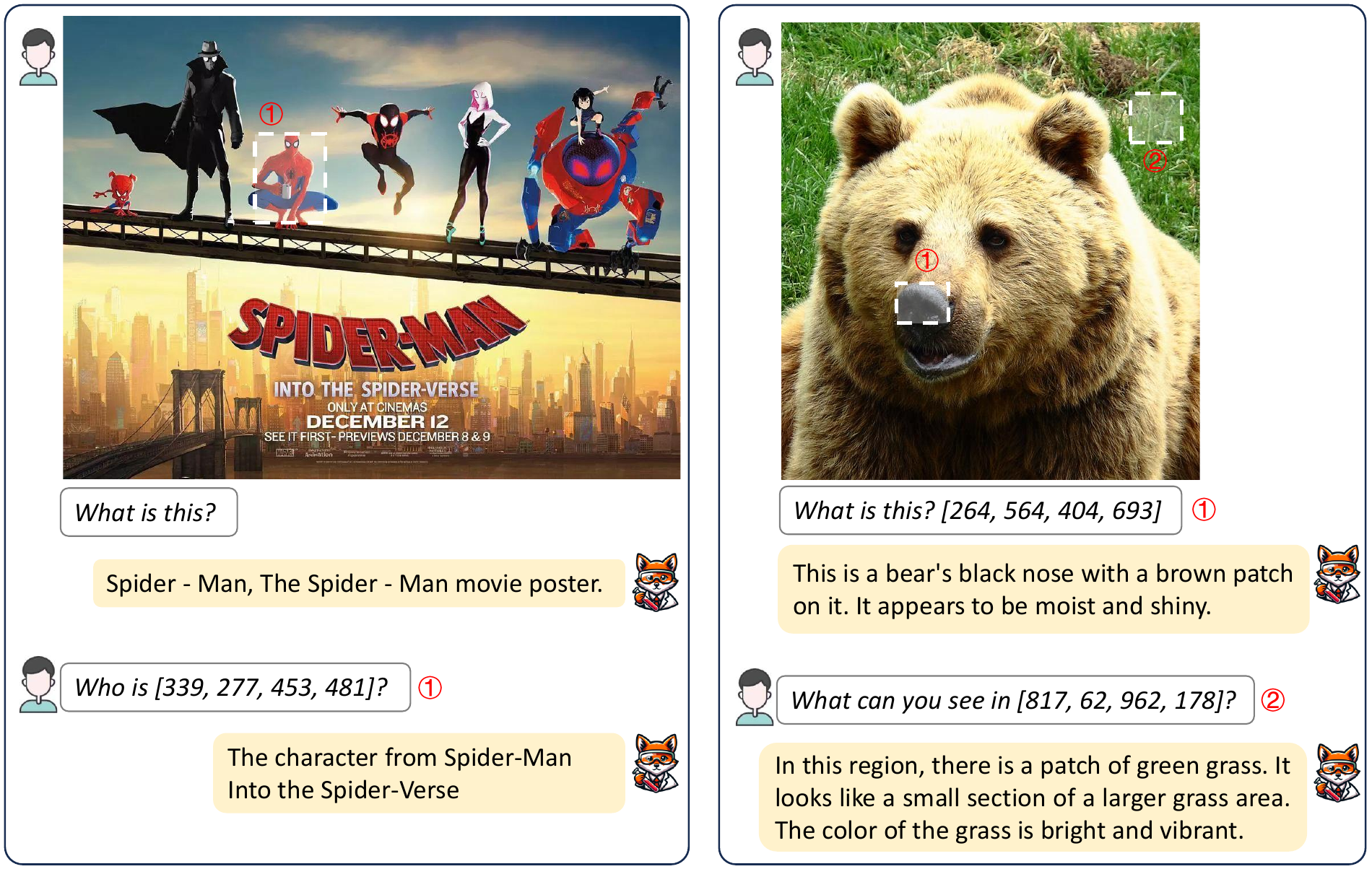}
\caption{Of course, Fox can yield interesting results in cartoon and natural scenes.}
\label{fig:append5}
\end{figure}

\newpage

{
\small
\bibliographystyle{splncs04}
\bibliography{egbib}
}

\end{document}